\documentclass{article}

\usepackage{multirow}

\usepackage{microtype}
\usepackage{graphicx}
\usepackage{subfigure}
\usepackage{booktabs} 

\usepackage{hyperref}

\usepackage[accepted]{icml2024}

\usepackage{amsmath}
\usepackage{amssymb}
\usepackage{mathtools}
\usepackage{amsthm}

\usepackage[capitalize,noabbrev]{cleveref}

\theoremstyle{plain}

\theoremstyle{definition}

\theoremstyle{remark}

\usepackage[textsize=tiny]{todonotes}

\icmltitlerunning{Distilling Morphology-Conditioned Hypernetworks for Efficient Universal Morphology Control}

\usepackage{xspace}
\newcommand{\model}{HyperDistill\xspace}

\begin{document}

\twocolumn[
\icmltitle{Distilling Morphology-Conditioned Hypernetworks\\for Efficient Universal Morphology Control}



\icmlsetsymbol{equal}{*}

\begin{icmlauthorlist}
\icmlauthor{Zheng Xiong}{oxford}
\icmlauthor{Risto Vuorio}{oxford}
\icmlauthor{Jacob Beck}{oxford}
\icmlauthor{Matthieu Zimmer}{huawei}
\icmlauthor{Kun Shao}{huawei}
\icmlauthor{Shimon Whiteson}{oxford}
\end{icmlauthorlist}

\icmlaffiliation{oxford}{Department of Computer Science, University of Oxford, Oxford, UK}
\icmlaffiliation{huawei}{Huawei Noah’s Ark Lab, London, UK}

\icmlcorrespondingauthor{Zheng Xiong}{zheng.xiong@cs.ox.ac.uk}

\icmlkeywords{Machine Learning, ICML}

\vskip 0.3in
]



\printAffiliationsAndNotice{}  

\begin{abstract}
Learning a universal policy across different robot morphologies can significantly improve learning efficiency and enable zero-shot generalization to unseen morphologies. 
However, learning a highly performant universal policy requires sophisticated architectures like transformers (TF) that have larger memory and computational cost than simpler multi-layer perceptrons (MLP). 
To achieve both good performance like TF and high efficiency like MLP at inference time, we propose \model, which consists of: (1) A morphology-conditioned hypernetwork (HN) that generates robot-wise MLP policies, and (2) A policy distillation approach that is essential for successful training of the HN. 
We show that on UNIMAL, a benchmark with hundreds of diverse morphologies, \model performs as well as a universal TF teacher policy on both training and unseen test robots, but reduces model size by 6-14 times, and computational cost by 67-160 times in different environments. 
Our analysis attributes the efficiency advantage of \model at inference time to knowledge decoupling, i.e., the ability to decouple inter-task and intra-task knowledge, a general principle that could also be applied to improve inference efficiency in other domains. The code is publicly available at \url{https://github.com/MasterXiong/Universal-Morphology-Control}. 
\end{abstract}

\section{Introduction}
\label{sec:intro}

Reinforcement learning (RL) for robotic control has made great progress in recent years \citep{levine2016end,kalashnikov2018scalable,andrychowicz2020learning,brohan2022rt}. 
However, generalization across robots remains a key challenge, as the policy trained for one robot transfers poorly to another robot with a different morphology, i.e., the topology graph of a robot and the hardware parameters of each body part. 
Furthermore, it is too sample inefficient to train a separate policy for each new robot from scratch. 

To tackle this challenge, universal morphology control aims to learn a universal policy to control different robots. 
By training on a set of diverse morphologies, this constitutes a form of multi-task RL \citep{vithayathil2020survey} where controlling each robot is a separate task. 
The learned policy is expected to not only improve learning efficiency on the training robots, but also enable zero-shot generalization to test robots with unseen morphologies. 

Training a multi-layer perceptron (MLP) policy is usually sufficient to achieve good performance on a single robot, but often generalizes poorly to other robots \citep{wang2018nervenet}. 
While it is feasible to train a multi-robot MLP policy for universal morphology control, it is significantly outperformed by graph neural networks (GNN) \citep{wang2018nervenet,pathak2019learning,huang2020one} and transformers (TF) \citep{kurin2020my,gupta2022metamorph} w.r.t.\ both training and generalization performance. 
Moreover, TF outperforms GNN by better modeling interactions between distant nodes in the morphology graph \citep{kurin2020my}, and has thus been adopted by most recent approaches \citep{dong2022low,furuta2022system,gupta2022metamorph,hong2022structureaware,trabucco2022anymorph,chen2023subequivariant,xiong2023universal}. 

However, TF has significantly higher memory, computation, and energy costs than MLP, all of which are key considerations when deploying the policy on real-world robots with constrained hardware  \citep{hutter2016anymal,zhao2021real,brohan2022rt,leal2023sara}. 
For example, a state-of-the-art TF-based method for universal morphology control \citep{xiong2023universal} requires about 40M FLOPs for a single step of inference on a robot with just 10 limbs, more than 100 times that of a single-robot MLP policy with similar performance, and this efficiency gap increases proportionally with the number of limbs. 
So a natural question arises: \emph{Can we learn a universal policy with the performance of TF but the inference efficiency of MLP?}

In this paper, we answer this question affirmatively. 
To get the best of both worlds, we introduce a hypernetwork (HN) \citep{ha2016hypernetworks}, i.e., a network that takes context features as input to generate the parameters of a base network. 
Our key intuition is that, compared to TF, HN can better decouple inter-task knowledge and intra-task knowledge via its hybrid architecture for more efficient inference, which we call the \emph{knowledge decoupling hypothesis}. 
Under our problem setting, the context-conditioned HN can provide sufficient model capacity to accommodate inter-robot knowledge, while the generated base network only needs to encode task-specific knowledge about the robot it controls. 
By contrast, TF encodes both kinds of knowledge with a single large model, which introduces high model redundancy when deployed on a specific robot. 

Specifically, suppose we train a set of single-robot MLP policies $ \left\{ \pi_k \right\}$, each with good performance on a different robot $k$. 
Motivated by the intuition that the optimal control policy of a robot critically depends on its morphology \citep{gupta2022metamorph,xiong2023universal}, we train an HN that takes the morphology context $c_k$ of robot $k$ as input to predict the corresponding MLP policy parameters $\pi_k$. 
For each robot, as the morphology context $c_k$ is constant, the HN-generated parameters are also fixed. 
Consequently, we only need to call it once before deployment to generate the base MLP, while the HN itself is not needed for policy execution. 
This yields a universal policy that works like an MLP at inference time but still has the potential to achieve good performance on both training and unseen test robots. 

While training such an HN policy via RL is a straightforward option, empirically we find that it is unstable and significantly underperforms a universal TF policy. 
Instead, we adopt a policy distillation (PD) approach by distilling a universal TF teacher policy into an HN student policy via behavior cloning (BC). 
While it is not hard to distill a student policy to match the teacher's performance on the training task(s) \citep{parisotto2015actor,rusu2015policy,czarnecki2019distilling}, a key challenge in our problem setting is to maintain the teacher policy's zero-shot generalization performance after distillation. 
In this paper, we identify several critical algorithmic choices in PD that influence the generalization performance of the student policy to unseen tasks: (1) The choice of the teacher(s), (2) Architecture alignment between the teacher and the student, (3) The number of tasks on which to collect training data for PD, and (4) Regularization in task space. 
We believe that these algorithmic choices could serve as general guidelines for improving task generalization of PD in other domains as well. 

We name our approach as \emph{\model} to highlight its two key components: (1) The HN architecture to achieve both good performance and high inference efficiency via knowledge decoupling, and (2) Training via policy distillation to successfully learn such an HN policy. 
We experiment on a challenging universal morphology control benchmark called UNIMAL \citep{gupta2021embodied}, which includes hundreds of diverse morphologies to evaluate both multi-robot training and zero-shot generalization. 
\model achieves performance similar to the universal TF teacher on both training and unseen test robots, while significantly reducing the model size by 6-14 times, and computational cost by 67-160 times in different environments at inference time. 
The experimental results further support our knowledge decoupling hypothesis, which could serve as a general principle to improve inference efficiency in other domains. 

\section{Background}
\subsection{Problem Formulation}
\label{subsec:formulation}
Learning a universal policy to control different morphologies can be seen as solving a contextual Markov Decision Process (CMDP) \citep{hallak2015contextual}, where the task context space $\mathcal{C}$ is defined over all possible morphology configurations. 
For robot (task) $k$, we use $\mathcal{S}_k$, $\mathcal{A}_k$, $T_k$ and $R_k$ to represent its state space, action space, transition function and reward function respectively. 

We assume that all the robots are drawn from a modular design space, i.e., each robot can be seen as a morphology tree over a set of basic nodes (limbs), and the node-level state and action space are homogeneous across different robots' limbs. 
Based on this assumption, we have $ \mathcal{S}_k = \mathcal{S}_k^1 \times \cdots \times \mathcal{S}_k^{N_k} $ and $ \mathcal{A}_k = \mathcal{A}_k^1 \times \cdots \times \mathcal{A}_k^{N_k}$, where $N_k$ is the number of limbs in robot $k$. 
The task context $c_k$ includes morphology information about the robot, consisting of node-wise context features $\{ c_k^i | i=1, \dots, N_k \}$ (see Appendix \ref{appendix:context-features} for more details), and a topology tree of the robot. 

We use $s_{k,t}, a_{k,t}, r_{k,t}$ to represent the state, action and reward at time step $t$ for robot $k$. 
The training objective is to learn a universal policy $\pi_\theta(a_{k,t}|s_{k,t}, c_k)$ to maximize the average return over a set of $K$ training robots, i.e., $ \max_{\theta} \left[ \frac{1}{K} \sum_{k=1}^K \sum_{t=0}^H r_{k,t} \right] $, where $H$ is the task horizon for all different robots. 
In addition to good training performance, we also expect the learned policy to generalize well on unseen test morphologies in a zero-shot manner. 

\subsection{Architectures for Universal Morphology Control}

To learn an MLP policy for the state and action space defined over a morphology tree, we first need some way to order the limbs, so that we can concatenate their node-wise states and actions into single vectors as the MLP's input and output. 
As the morphology of a robot has a tree structure, the limbs in each robot are usually ordered by some tree traversal methods like depth-first search \citep{gupta2022metamorph}. 

When learning a multi-robot MLP policy, to handle the variable number of limbs across different robots, it is common to assume a maximal limb number $N_{\text{max}}$, and zero-pad the state and action vectors to this maximal length so that the state and action dimensions are aligned across different robots. 
As the universal policy also conditions on the morphology context, the context features $c_k^i$ are usually concatenated with the state features $s_{k,t}^i$ as node-wise input \citep{gupta2022metamorph}. 
So the final input to the multi-robot MLP policy is $x=\left[ x_1, x_2, \dots, x_{N_k}, \vec{0}, \dots, \vec{0} \right]$, where $x_i=[s_i, c_i]$, and we omit the robot index $k$ and time index $t$ thereafter for simplicity of notation. 
For the input layer of the MLP, we have $h^{(0)}=\sigma \left( W^{\text{in}} x + b^{\text{in}} \right) = \sigma \left( \sum_{i=1}^{N_k} W^{\text{in}}_i x_i + b^{\text{in}} \right) $, where $\sigma$ is the activation function, $W_i^{\text{in}}$ is the subset of the weight matrix that encodes the input features of node $i$. 
For the $l$-th hidden layer, we have $h^{(l+1)} = \sigma \left( W^{(l)}h^{(l)} + b^{(l)} \right)$. 
For the output layer, we have $a = W^{\text{out}} h^{(L)} + b^{\text{out}}$, which can be decomposed into node-wise action $a_i = W_i^{\text{out}} h^{(L)} + b_i^{\text{out}}$. 

Unlike MLP, GNN \citep{wu2020comprehensive} and TF \citep{vaswani2017attention} can naturally handle variable numbers of homogeneous elements. 
GNN can be directly applied to different morphology graphs, while TF treats each limb as a token. 
Unlike in many domains where tokens are processed sequentially, in universal morphology control, TF processes different nodes in parallel to model their spatial interactions. 
A typical TF architecture for universal morphology control consists of a linear embedding layer that embeds node-wise input features, multiple attention blocks that model limb interactions, and an MLP decoder that maps the node-wise embedding outputed by the attention blocks to node-wise actions \citep{gupta2022metamorph,xiong2023universal}. 

\subsection{Hypernetworks}

A hypernetwork \citep{ha2016hypernetworks} is a network that generates the parameters of a base network conditioned on some context $c$. 
We can decompose an HN into a context encoder $f$ and several output heads (parameterized as linear layers) that take the context embedding $e=f(c)$ as input, and output parameters in the base network. 
For example, to generate the parameters of a linear layer $y=Wx+b$, the HN needs two output heads to generate $W$ and $b$ respectively, i.e., $W=\text{HN}_W(e)$, $b=\text{HN}_b(e)$. 
If $W$ is a $M \times N$ matrix, and the context embedding dimension is $E$, then $\text{HN}_W$ is a linear layer with input dimension $E$ and output dimension $M \times N$, while $\text{HN}_b$ is a linear layer with input dimension $E$ and output dimension $N$. 

\section{HyperDistill}
This section introduces \model, which achieves both good performance and high efficiency at inference time. 
In Section \ref{subsec:limitation}, we analyze the limitations of existing methods for universal morphology control, which motivates us to introduce HN as a solution in Section \ref{subsec:architecture}. 
In Section \ref{subsec:PD}, we discuss why and how we train the HN via policy distillation, and analyze some of its critical algorithmic designs.

\subsection{Limitations of Existing Architectures}
\label{subsec:limitation}
We analyze the limitations of two representative architectures for universal morphology control: TF, which achieves SOTA performance but is expensive to run at inference time, and MLP, which runs efficiently and achieves good performance on each single robot but performs poorly in the multi-robot setting. 
We first highlight a knowledge decoupling issue that exists in both TF and MLP, then discuss the lack of order-invariance in MLP. 

\paragraph{Knowledge Decoupling}
Intuitively, knowledge learned by a universal policy can be decomposed into two parts: inter-task knowledge for generalization across robots, and intra-task knowledge about how to control a specific robot. 
Since both TF and MLP use a single set of parameters to encode both kinds of knowledge, they cannot decouple them from each other. 
Consequently, at inference time, there may be a lot of redundant knowledge in the policy that is irrelevant to solving the current task, which harms efficiency. 
It also explains why MLP can perform well on a single robot but not under the multi-robot setting, as the model capacity of a compact MLP is sufficient to accommodate task-specific knowledge of a single robot, but not enough to encode both inter-task and intra-task knowledge under a multi-robot setting. 
The same conclusion also holds for TF when compressing a large TF into a smaller one, as we confirm experimentally in Section \ref{sec:experiments}. 
In summary, due to lack of knowledge decoupling, TF and MLP have to trade off between performance and efficiency at inference time. 

\paragraph{Order-Invariance Issue with MLP}
A further issue with MLP is that it is not invariant to the order of limbs across robots. 
TF is order-invariant, as all the limbs across different robots share the same set of parameters. 
By contrast, in a multi-robot MLP, only the limbs with the same index across different robots share the same subset of parameters in the input and output layers, so how we order the limbs influences the policy output. 
As we do not have a consistent way to order the limbs across different morphologies, multi-robot MLP can overfit to spurious patterns in the manually chosen limb indexing method, harming generalization.

\subsection{\model Architecture}
\label{subsec:architecture}
To tackle the limitations of TF and MLP, we introduce an HN that takes morphology context as input to generate an MLP base policy for each robot. 
First, it supports knowledge decoupling, as the morphology-conditioned HN encodes inter-task knowledge, while the base MLP encodes intra-task knowledge. 
The expensive HN is called only once to generate a task-specific base policy for each new task, while the much smaller base network is sufficient to efficiently solve the task. 
Second, the base MLP generated by the HN is order-invariant, as the parameters associated with each limb no longer depend on the limb index, but only condition on the limb's context representation. 

\begin{figure}[t]
    \centering
    \includegraphics[width=\columnwidth]{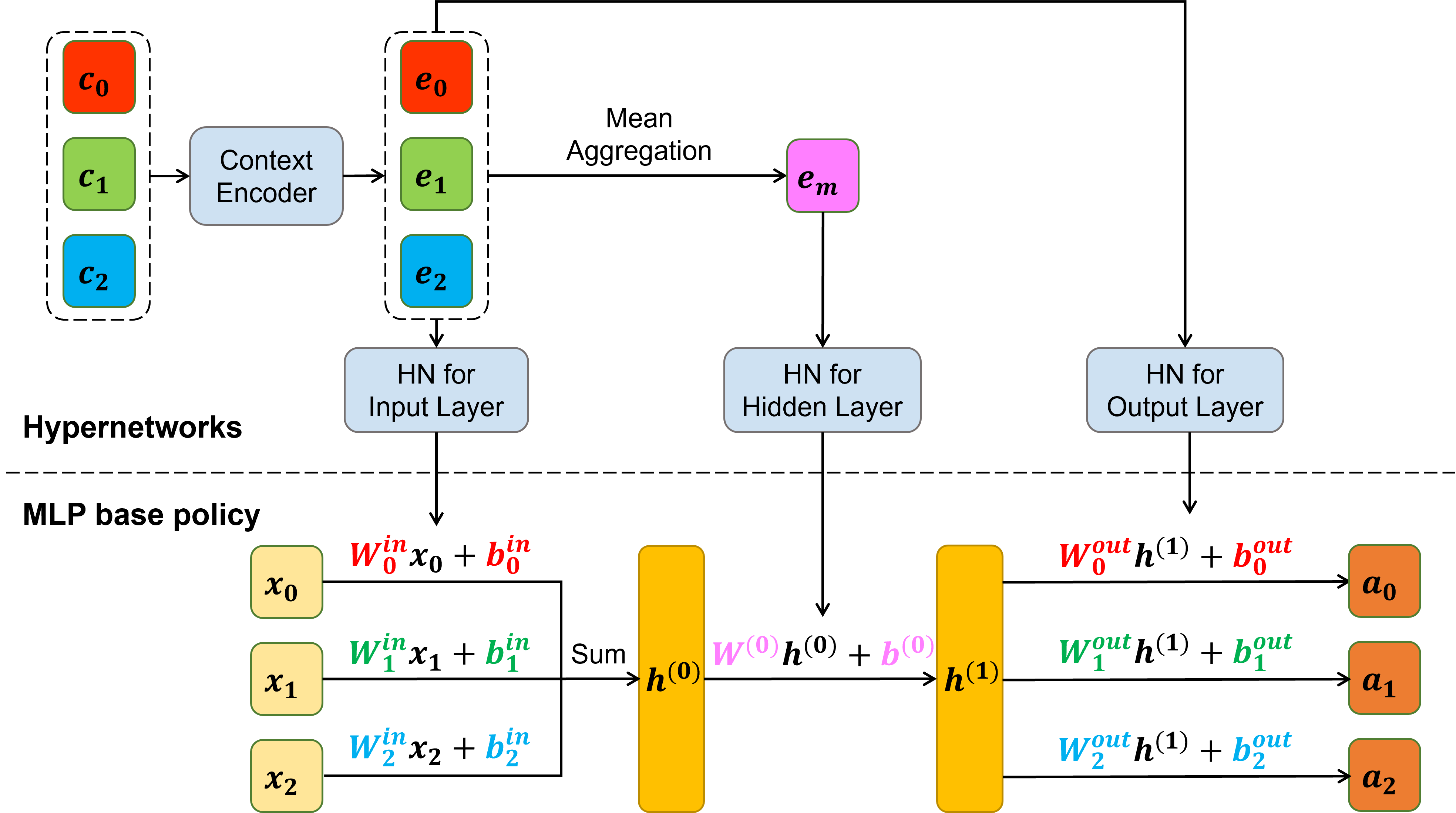}
    \caption{The architecture of \model. Different colors highlight the correspondence between the parameters in the base network and the context embedding they condition on via HN. We only show one hidden layer in the base MLP for ease of illustration. More hidden layers can be easily added in a similar way.}
    \label{fig:framework}
\end{figure}
The overall architecture is shown in Figure \ref{fig:framework}. 
Unlike prior work in HN, a unique challenge in applying HN to universal morphology control is that the input and output dimensions of the MLP base network vary across robots due to the variable number of limbs.
We tackle this challenge by generating limb-wise parameters for the input and output layers, as the subset of parameters corresponding to each limb still has fixed dimension based on the modular space assumption in Section \ref{subsec:formulation}. 
Specifically, for the input layer, we have $ h^{(0)}=\sigma \left( \sum_i W^{\text{in}}_i x_i+b^{\text{in}}_i \right) $, 
where $ W_i^{\text{in}} = \text{HN}_W^{\text{in}}(e_i) $, $ b^{\text{in}}_i = \text{HN}_b^{\text{in}}(e_i) $, and $e_i = f(c_i)$ is the context embedding of limb $i$ generated by the context encoder $f$. 
As the whole policy conditions on the morphology context through HN, we no longer need to concatenate context features $c_i$ to the network input, so we have $x_i = s_i$. 
Similarly, for the output layer, we have $ a_i = W_i^{\text{out}} h^{(L)} + b_i^{\text{out}}$, where $W_i^{\text{out}} = \text{HN}_W^{\text{out}}(e_i)$, and $ b_i^{\text{out}} = \text{HN}_b^{\text{out}}(e_i)$. 
For the hidden layers, we first aggregate the context embedding of different limbs using the mean to get a context embedding for the whole morphology as $e_m = \frac{1}{N}\sum_i e_i$, then pass $e_m$ through HN output heads to generate the hidden layer parameters, i.e., $ h^{(l+1)} = \sigma \left( W^{(l)}h^{(l)} + b^{(l)} \right) $, where $ W^{(l)} = \text{HN}_W^{(l)}(e_m) $, and $ b^{(l)} = \text{HN}_b^{(l)}(e_m) $.

\paragraph{Context Representation Learning}
The quality of the HN-generated policy critically depends on the quality of the learned context embedding, e.g., if $e_i$ of different limbs are too similar, the policy will generate similar actions across different limbs, which is unlikely to be a good policy. 

The context representation should be both discriminative to encode the diverse behaviors of different limbs, and generalizable to unseen robots. 
We adopt two approaches to achieve this goal. 
First, we apply some simple transformations to the context features $c_i$ to make them more discriminative across limbs (see Appendix \ref{appendix:context-features}). 
Second, we use a TF as the context encoder to enrich each limb's context representation by interacting with other limbs in the robot, e.g., if two limbs in two different morphologies have the same hardware configurations, then we cannot tell them apart based on their own context features alone, but the TF context encoder can learn a distinguishable representation by further encoding morphology information. 
Since the TF context encoder is also a part of the HN, it is not needed at inference time, unlike a TF policy that uses TF as the controller. 

\subsection{\model Training via Policy Distillation}
\label{subsec:PD}

In principle, we can train a universal HN policy from scratch via RL. 
However, empirically we find that the training process is unstable and the learned policy significantly underperforms a TF policy, possibly because both HN and RL are known to be unstable during learning, and combining them together further exacerbates the optimization challenges. 

Instead, we adopt policy distillation \citep{parisotto2015actor,rusu2015policy} by first training a universal TF policy, then distilling it into an HN policy via behavior cloning, which replaces RL training with more stable supervised learning to alleviate the optimization challenge. 
Moreover, as BC empirically requires much less time to train than RL, we can reuse the same pre-trained teacher policy for more efficient evaluation of different algorithmic choices. 

Given a set of training robots and a universal TF policy trained on them as the teacher $\pi^T$, we first collect expert trajectories on each training robot with $\pi^T$ to generate a training dataset $\mathcal{B}^T$ for PD. 
Then we train an HN student policy $\pi$ by minimizing the KL-divergence between the action distributions generated by the teacher and student on transitions randomly sampled from the buffer: $$ L_{\pi} = \mathbb{E}_{s, c \sim \mathcal{B}^T} \left[ KL(\pi^T(a|s, c)||\pi(a|s, c) \right]. $$
While more sophisticated loss functions \citep{czarnecki2019distilling} have been proposed to facilitate PD by collecting online data with the student, empirically we find that this simple BC loss is sufficient to learn a student policy that matches the teacher's performance on the training robots, without having to collect further samples with the student. 

However, we also want the student to zero-shot generalize as well as the teacher on unseen test robots. 
While several prior works evaluate zero-shot generalization of a student policy to unseen tasks in different problem settings \citep{chen2022system,furuta2022system,wan2023unidexgrasp++}, none of them systematically investigate how different algorithmic choices in PD influence the student's generalization performance. 
In this paper, we highlight four key factors that may influence the generalization gap between teach and student in PD. 

\paragraph{The Choice of PD Teacher(s)}
\label{subsec:PD_teacher_choice}
Theoretically, we can either train a universal TF teacher on multiple robots, or train multiple single-robot MLP policies on different morphologies as the teachers \citep{furuta2022system}. 
While the average performance of the single-robot MLPs is similar to or even better than that of a universal TF on the training robots \citep{xiong2023universal}, we hypothesize that the student policy distilled from a universal teacher policy generalizes better to unseen robots. 
Intuitively, as the single-robot teacher policies are trained via RL separately, they can be dramatically different from each other, which may exacerbate the discontinuity of the distilled policy in the parameter space and harm generalization. 

\paragraph{Student-Teacher Architecture Alignment}
We hypothesize that the generalization gap between teacher and student is smaller when their neural architectures are more aligned \citep{hao2023one}. 
Intuitively, the inductive bias introduced by a more aligned architecture helps the student extrapolate to unseen task context and state in a more consistent way with the teacher, which helps reduce the generalization gap. 

However, as we are trying to distill from a TF to an HN, there is an unavoidable mismatch between the teacher and student architectures. 
The two algorithmic choices discussed next may help compensate for the generalization gap caused by this architecture misalignment. 

\paragraph{Number of Tasks for Distillation Training}
To better align the teacher and student policies' behaviors on unseen tasks, a natural idea is to train the student policy on the teacher's demonstrations collected from more tasks, so that the PD training data better covers the task space.
This is different from training the teacher policy on more tasks, which is an effective but orthogonal way to improve task generalization. 
Our approach does not require modifying the teacher's training process. 
Instead, it simply requires collecting data from more tasks with the pre-trained universal teacher, which introduces little computational overhead. 
For clarity, we use ``training robots'' to denote the robots on which we train the teacher policy, and  ``PD robots'' to denote the robots on which we collect expert trajectories for PD. 
Advanced methods like curriculum learning \citep{dennis2020emergent,narvekar2020curriculum,wan2023unidexgrasp++} could be utilized to get a more robust task distribution to further mitigate generalization gap, which we leave for future work. 

\paragraph{Regularization in Task Space}
Regularization is a common approach to reduce overfitting and improve generalization \citep{cobbe2019quantifying}. 
We consider it to be especially important for \model, as HN may be more prone to overfitting than other architectures. 
In prior work, morphology context is usually just used as an additional policy input \citep{gupta2022metamorph}, while in \model, it is used as HN input to generate the whole control policy. 
As we only have a few hundred different robots for PD, which is much less than the number of parameters in the HN, there is a high chance of overfitting. 
To reduce overfitting, we apply dropout to the context embedding $e_i$ and $e_m$. 
This can be seen as regularization in task space, which encourages the HN to learn an ensemble of different MLP policies that can all work well on the same robot. 
It can also be seen as domain randomization \citep{tobin2017domain} by making the generated policy more robust to changes in morphology context. 
We leave it for future work to investigate other regularization methods like weight decay. 

\section{Experiments}
\label{sec:experiments}
\begin{figure*}[t]
    \centering
    \includegraphics[width=\textwidth]{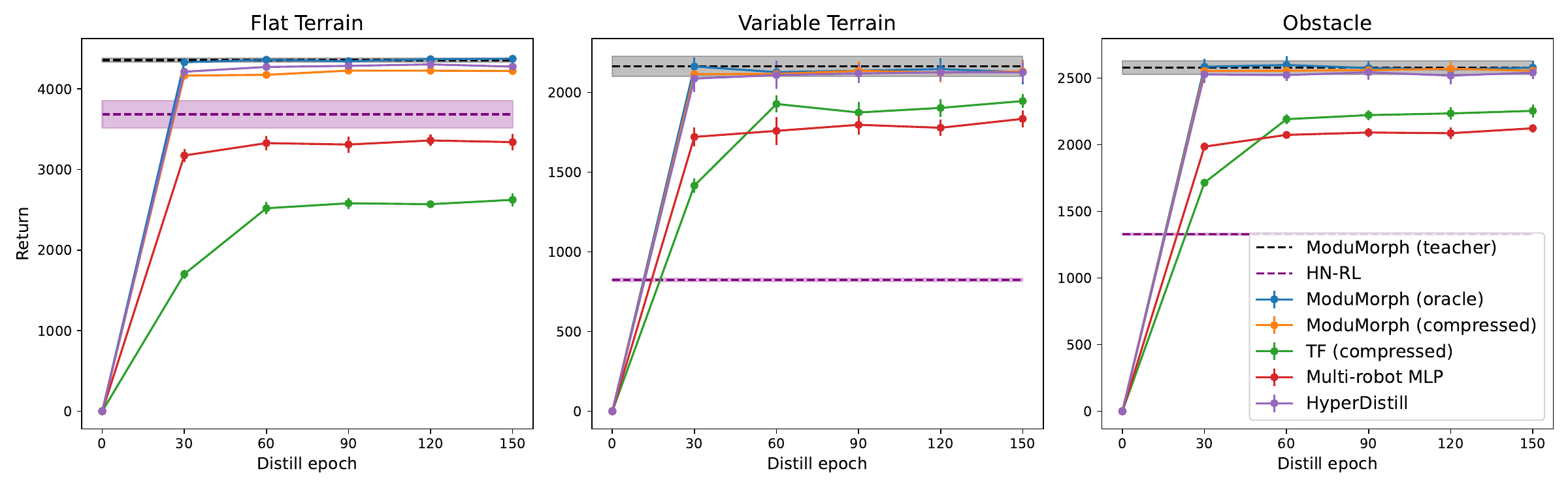}
    \caption{The performance of different methods on the \emph{training} robots in each environment.}
    \label{fig:distill-train}
\end{figure*}
Our experiments aim to answer the following questions: 

(1) Can \model achieve good performance on both training and unseen test robots? How does it compare to other methods w.r.t.\ performance and efficiency at inference time? (Section \ref{subsec:main_results})

(2) How do different algorithmic and architecture choices in \model influence its training and generalization performance? (Section \ref{subsec:ablation_distillation})

\subsection{Experimental Setup}

We experiment on the UNIMAL benchmark \citep{gupta2021embodied} built upon the Mujoco simulator \citep{todorov2012mujoco}, which includes 100 training robots and 100 test robots with diverse morphologies, while new morphologies can be easily generated via mutation operations supported by the UNIMAL design space. 
Following the setup of \citet{gupta2022metamorph}, we consider three different environments with increasing difficulties: 
(1) Flat terrain (FT): maximize locomotion distance on a flat floor; 
(2) Variable terrain (VT): maximize locomotion distance on a variable terrain randomly reset for each episode. 
(3) Obstacle: maximize locomotion distance on a flat terrain while avoiding randomly positioned obstacles. 

\paragraph{Teacher Policy}
For each environment, we train a universal TF policy on the 100 training robots as the teacher policy. 
We adopt ModuMorph \citep{xiong2023universal} as the TF teacher, as it achieves SOTA performance on the UNIMAL benchmark. 
ModuMorph differs from a standard TF architecture in two ways: 
(1) In the attention blocks, it computes a fixed attention matrix solely conditioned on the morphology context, which performs better than computing dynamic attention weights based on limb observations. 
(2) The embedding layer and the MLP decoder are generated by HNs in ModuMorph. 
Although ModuMorph and \model both use HNs, their motivations are significantly different. 
ModuMorph adopts HNs to better model the diverse behaviors across limbs, but is still a large TF-based model with high inference costs. 
By contrast, \model utilizes HNs to enable knowledge decoupling for efficient inference. 
This novel perspective on the role of HNs is a key contribution of our method. 
Furthermore, as we show in Section \ref{subsec:main_results}, a TF-based model can be compressed to a much smaller size without much performance loss, but only when equipped with these HN-generated layers, while a standard TF cannot. 

\paragraph{Data Collection for Distillation}
To collect data for policy distillation, we first generate an augmented robot set of 1,000 PD robots by mutating the 100 training robots. 
See Appendix \ref{appendix:PD-robot-generation} for how we do the mutation. 
Then we collect 8,000 transitions from each PD robot, forming a multi-robot dataset with 8M transitions for policy distillation. 

\paragraph{Baselines}
We compare \model with the following architectures as the student policy: 
\textbf{(1) ModuMorph (oracle): }A ModuMorph student with the same architecture as the teacher. It serves as a performance upper bound on how well the student can perform without architecture misalignment. 
\textbf{(2) TF (compressed): }A standard TF with a similar number of parameters as the base MLP in \model. It is used to validate that standard TF cannot achieve both good performance and high efficiency at the same time like \model. 
\textbf{(3) ModuMorph (compressed): } A ModuMorph student with a similar number of parameters as the base MLP in \model. As it adopts HNs to generate some layers, we expect it to have better knowledge decoupling ability than standard TF, but may still be less efficient than \model due to the attention blocks. 
\textbf{(4) Multi-robot MLP: }A multi-robot MLP with the same architecture as the base MLP in \model, which is used to validate the advantages of \model over MLP as discussed in Section \ref{subsec:architecture}. 
Finally, we also compare with training an HN policy via RL (\textbf{HN-RL}) to show the importance of distillation. 

\paragraph{Distillation Setup}
The distillation process runs for 150 epochs, with a mini batch size of 5120. 
We use Adam with a learning rate of 0.0003, and clip the gradient norm to 0.5. 
We run three random seeds for each method in each environment, and report the average performance with standard error. 
For \model, we apply dropout to context embedding with $p=0.1$. 
See Appendix \ref{appendix:model-size} for the size of each student model in each environment. 

\subsection{Main Results}
\label{subsec:main_results}
\begin{figure*}[t]
    \centering
    \includegraphics[width=\textwidth]{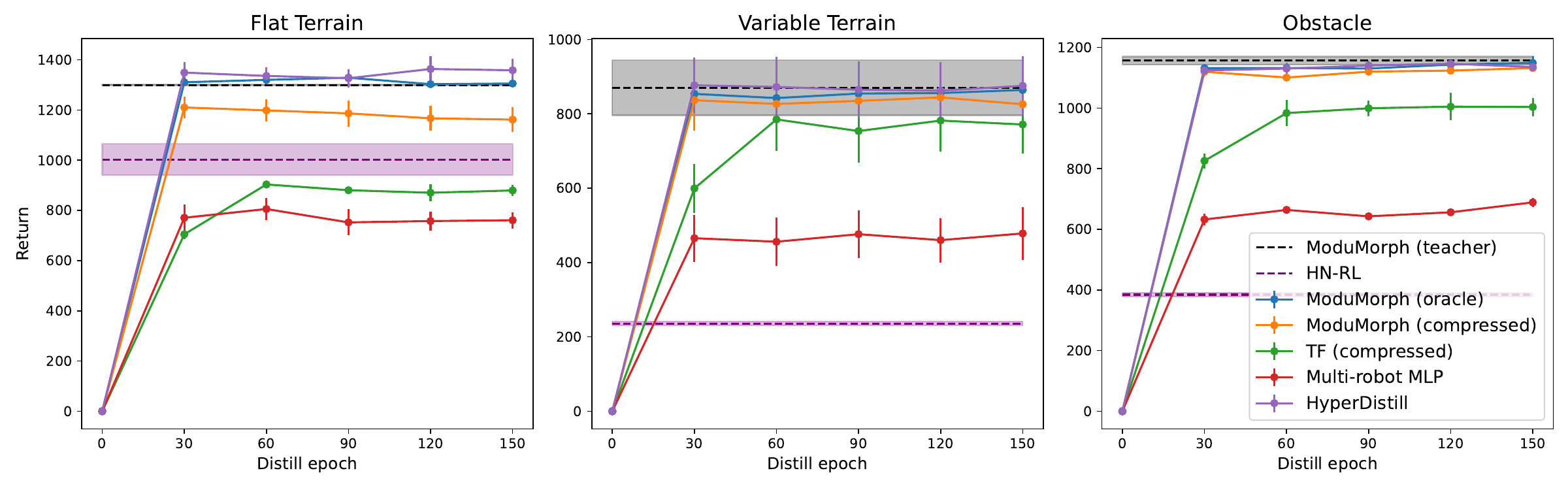}
    \caption{The performance of different methods on the \emph{test} robots in each environment.}
    \label{fig:distill-test}
\end{figure*}
Figures \ref{fig:distill-train} and \ref{fig:distill-test} illustrate how different methods perform on the training and test robots during the distillation process. 
Table \ref{tab:distill-result} compares the model size and FLOPs of different methods at inference time. 
\begin{table}[h]
    \centering
    \resizebox{\columnwidth}{!}{
    \begin{tabular}{clcccc}
        \toprule
        Task & \multicolumn{1}{c}{Method} & \multicolumn{2}{c}{Model size} & \multicolumn{2}{c}{FLOPs} \\
         & & Abs. & Rel. & Abs. & Rel. \\
        \midrule
        \multirow{4}{*}{FT} & ModuMorph (oracle) & 1.73 M & 14.0 & 39.86 M & 160.8 \\
         & TF (compressed) & 0.14 M & 1.1 & 3.39 M & 13.7 \\
         & ModuMorph (compressed) & 0.15 M & 1.2 & 1.78 M & 7.2 \\
         & Multi-robot MLP & 0.23 M & 1.9 & 0.46 M & 1.9 \\
         & \textbf{\model} & \textbf{0.12 M} & \textbf{1} & \textbf{0.25 M} & \textbf{1} \\
        \hline
        \multirow{4}{*}{VT} & ModuMorph (oracle) & 1.97 M & 6.6 & 40.33 M & 67.2 \\
         & TF (compressed) & 0.31 M & 1.0 & 5.51 M & 9.2 \\
         & ModuMorph (compressed) & 0.39 M & 1.3 & 2.26 M & 3.8 \\
         & Multi-robot MLP & 0.41 M & 1.4 & 0.82 M & 1.4 \\
         & \textbf{\model} & \textbf{0.30 M} & \textbf{1} & \textbf{0.60 M} & \textbf{1} \\
        \hline
        \multirow{4}{*}{Obstacle} & ModuMorph (oracle) & 2.02 M & 6.7 & 40.43 M & 67.4 \\
         & TF (compressed) & 0.32 M & 1.0 & 5.61 M & 9.3 \\
         & ModuMorph (compressed) & 0.44 M & 1.5 & 2.35 M & 3.9 \\
         & Multi-robot MLP & 0.41 M & 1.4 & 0.82 M & 1.4 \\  
         & \textbf{\model} & \textbf{0.30 M} & \textbf{1} & \textbf{0.60 M} & \textbf{1} \\
        \bottomrule
    \end{tabular}
    }
    \caption{Model size and FLOPs of different methods \emph{at inference time}. Abs. denotes the absolute value, and Rel. denotes the relative value w.r.t.\ \model. See Appendix \ref{appendix:flops-computation} for how we compute the FLOPs, and \ref{appendix:efficiency-analysis} for more analysis on the results in this table.}
    \label{tab:distill-result}
\end{table}

\model achieves performance similar to ModuMorph (oracle), matching the teacher's performance on both the training and test robots in all the three environments, while reducing model size by 6-14 times, FLOPs by 67-160 times in different environments. 

TF (compressed) cannot match the performance of \model in all the three environments, as standard TF needs to trade off between performance and efficiency due to a lack of knowledge decoupling. 

As expected, ModuMorph (compressed) consistently outperforms TF (compressed), as it generates some layers of the TF via HNs, which enables better knowledge decoupling. 
However, it still lags behind \model w.r.t.\ both generalization performance and efficiency, as the knowledge encoded in the attention blocks still cannot be decoupled. 
This result further indicates that, in contrast to previous work \citep{kurin2020my,gupta2022metamorph,xiong2023universal}, we may not need complicated attention modules to achieve good performance for universal morphology control. 
In addition, the performance gap between ModuMorph (compressed) and \model is larger in FT than in VT and Obstacle, possibly because for the latter two environments, ModuMorph has more layers in the HN-generated decoder, which makes it more similar to \model. 

\model also significantly outperforms multi-robot MLP, which validates the importance of the HN. 
Moreover, the performance gap between multi-robot MLP and other methods is much larger on the test robots than on the training ones, which may be overfitting due to the order-invariance issue of MLP discussed in Section \ref{subsec:limitation}. 

HN-RL performs poorly on both training and test robots, which reflects the optimization difficulty of combining HN and RL, and validates the importance of training via PD. 

\subsection{Ablation Studies}
\label{subsec:ablation_distillation}
In this section, we investigate how the algorithmic choices in PD influence generalization performance of the student. 
Due to space limitations, see Appendix \ref{appendix:context-representation-ablation} for ablations on architecture choices in \model. 
To save computation, we run PD for only 50 epochs as most methods can already converge within this budget, and experiment only in the FT experiment unless otherwise stated. 

\paragraph{The Choice of PD Teacher(s)}
We train (1) A universal TF teacher on all the training robots; (2) A set of single-robot MLP teachers on each training robot. Then we collect 80,000 transitions from each training robot with the teacher(s) to get $\mathcal{B}^T$. 
We experiment with both \model and TF (oracle) to ensure that the PD choice applies to different student architectures. 
\begin{figure}[t]
    \centering
    \includegraphics[width=\columnwidth]{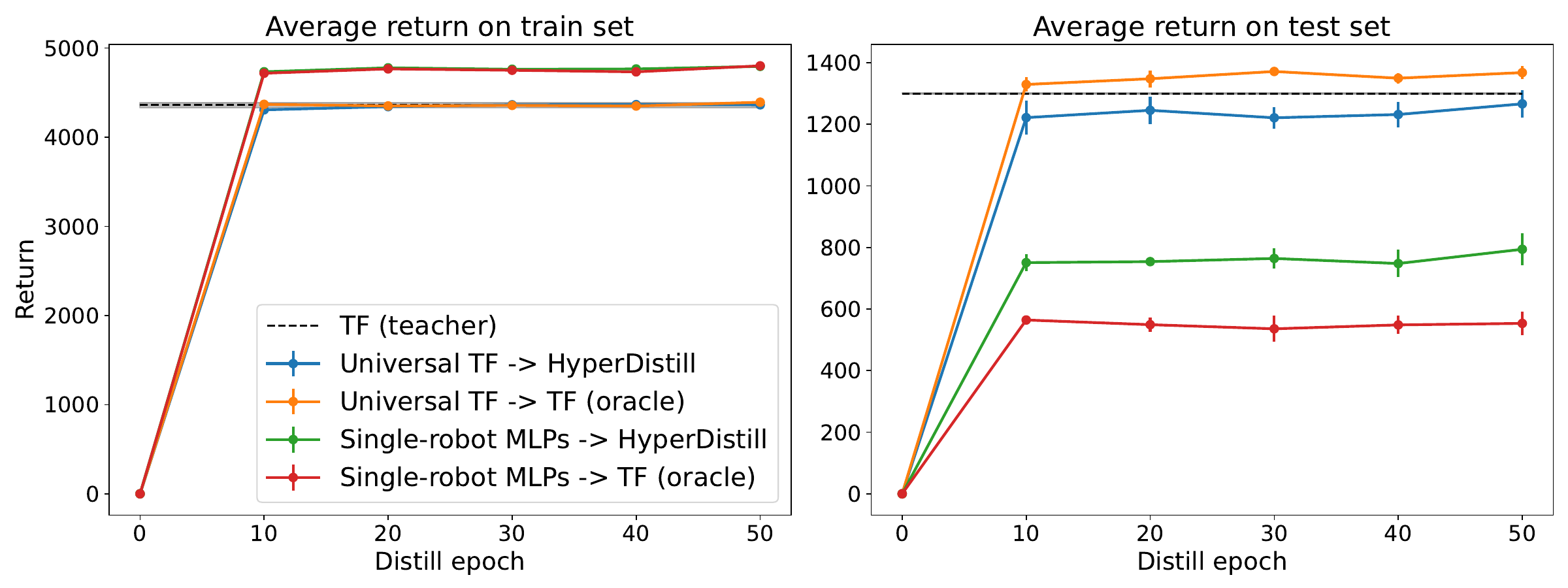}
    \caption{The student's learning curves under different teacher choices. In the figure legend, ``X $\rightarrow$ Y'' means that we distill from teacher X into student Y.}
    \label{fig:PD_teacher_choice}
\end{figure}
Figure \ref{fig:PD_teacher_choice} shows how the student policies perform on the training and test robots with different teacher choices. 
As expected, when using single-robot MLP teachers, although the student policies perform well on the training robots, they generalize much worse than the students distilled from a universal TF teacher, which validates our hypothesis in Section \ref{subsec:PD}. 
\begin{figure}[t]
    \centering
    \includegraphics[width=\columnwidth]{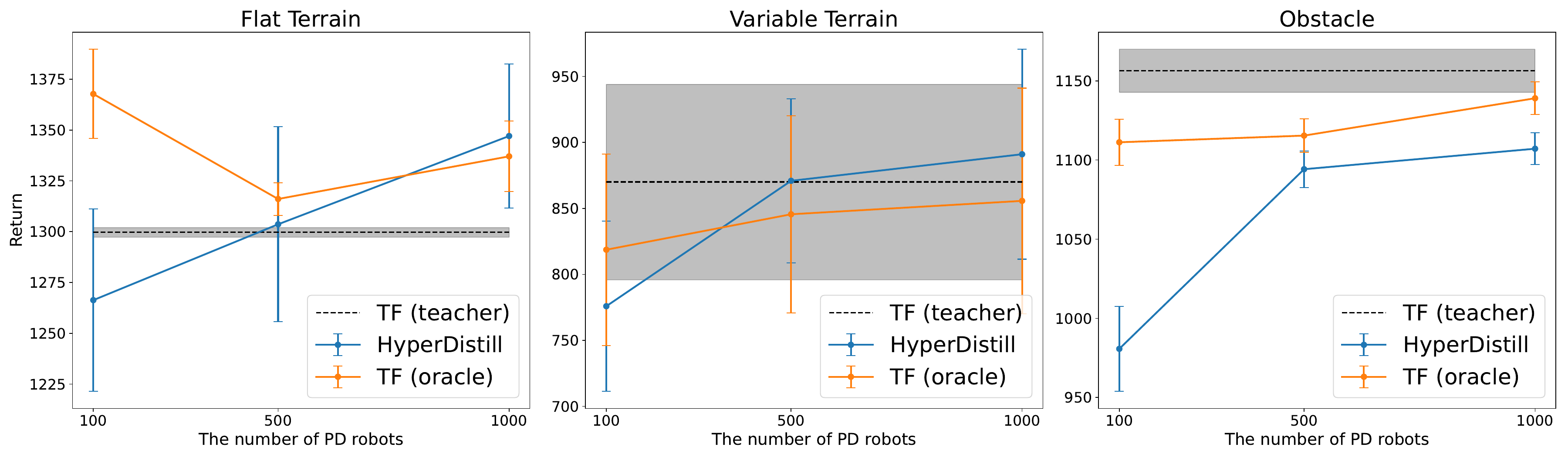}
    \caption{Final generalization performance of \model and TF (oracle) w.r.t.\ the number of PD robots in different environments.}
    \label{fig:PD_robot_num}
\end{figure}

\paragraph{Student-Teacher Architecture Alignment}
We reexamine the results in Figure \ref{fig:PD_teacher_choice} from a different perspective. 
When using a universal TF teacher, the TF student has a more aligned architecture and achieves better generalization performance than \model. 
When using single-robot MLP teachers, while both students generalize much worse, \model has a more aligned architecture (as its base MLP has the same architecture as the teachers), and generalizes better than the TF student. 
Moreover, for the same teacher, both students achieve similar performance on the training robots regardless of their architectures, indicating that the generalization gap is not caused by the difference in model capacity of different student models, but is more likely the consequence of architecture misalignment. 

\paragraph{Number of Distillation Training Tasks}
To validate our hypothesis that collecting distillation data from more robots can improve the student's task generalization, we experiment with 100, 500, and 1,000 PD robots. 
For a fair comparison, the number of transitions collected from each robot decreases proportionally so that the total data size remains unchanged. 
As shown in Figure \ref{fig:PD_robot_num}, \model's generalization performance increases by 6\%, 15\% and 13\% in the three environments as the number of PD robots increases from 100 to 1,000. 
There is no significant improvement in the TF student's generalization performance due to a ceiling effect and its better aligned architecture with the TF teacher. 

\paragraph{Regularization in Task Space}
\begin{figure}[t]
    \centering
    \includegraphics[width=\columnwidth]{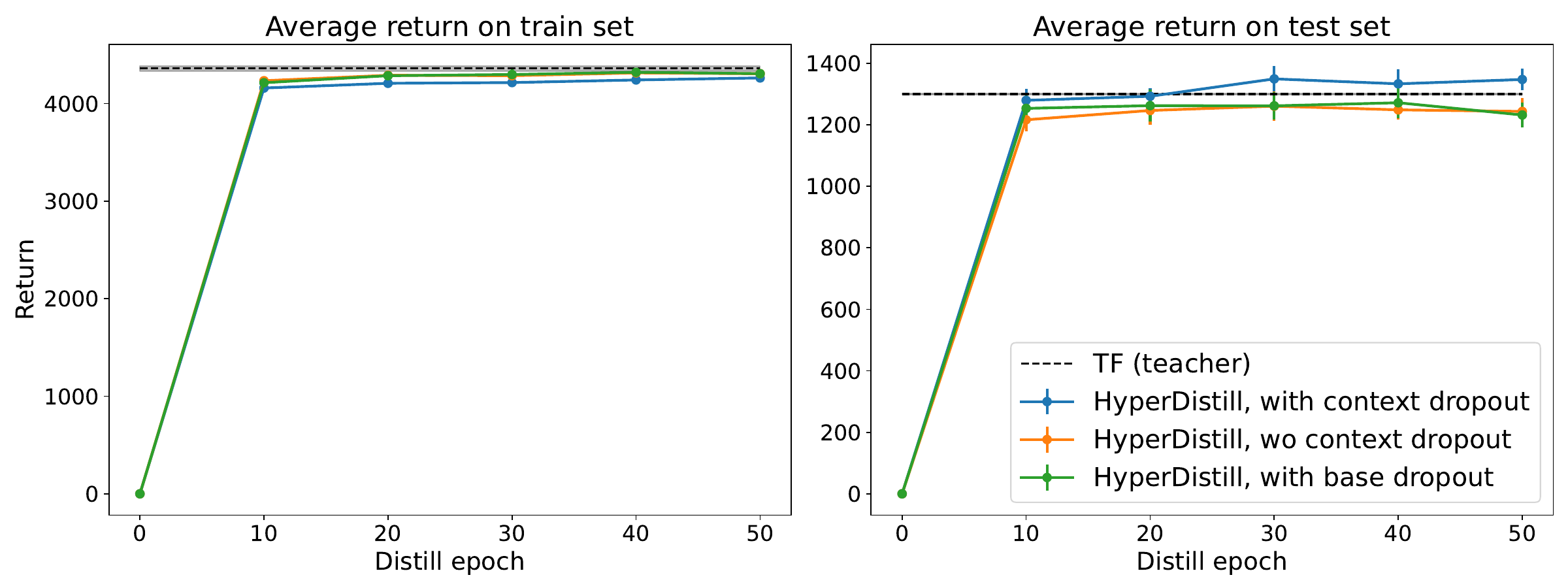}
    \caption{Learning curves of \model in FT with different ways of incorporating dropout regularization.}
    \label{fig:PD_dropout}
\end{figure}
As shown in Figure \ref{fig:PD_dropout}, applying dropout to the context embedding improves generalization performance by 8.5\%, while applying dropout to the base MLP does not provide significant improvement, which agrees with our intuition that regularization may be more important in task space than in state space. 

\section{Discussion}
We reexamine our knowledge decoupling hypothesis to see if it is supported by experimental results. 
First, \model achieves similar performance as TF but runs much more efficiently at inference time, which validates the efficiency benefits provided by knowledge decoupling. 
Second, as expected, due to the lack of knowledge decoupling, when we compress a universal TF policy, the compressed TF does not have enough model capacity to accommodate both inter-task and intra-task knowledge, as evidenced by its worse performance in our experiments. 
The knowledge decoupling hypothesis further suggests that if we compress a universal TF into a single-robot TF, it should outperform a compressed universal TF on that specific robot, as we relieve it of the burden of distilling inter-task knowledge. 
To validate this idea, we conduct a proof-of-concept experiment in the Obstacle environment. 
We randomly sample 10 training robots and distill the universal TF teacher into a single-robot TF with the same architecture as TF (compressed) for each robot. 
As expected, the compressed single-robot TFs achieve an average return of 2345, outperforming the compressed universal TF with an average return of 2115, but at the cost of losing the generalization ability to other robots. 

In summary, TF needs to trade off between performance and efficiency at inference time, while \model can enjoy both thanks to knowledge decoupling. 
It resembles related ideas explored in other fields, such as mixtures-of-experts \citep{riquelme2021scaling,shen2023scaling} that learn a large model with high capacity while sparsely activating sub-modules for different tasks to improve inference efficiency. 
There is also evidence from neuroscience that although the human brain contains billions of neurons, it is still power-efficient because it activates only a small fraction of neurons at a time to solve a given task \citep{barth2012experimental}. 
Consequently, we believe that our knowledge decoupling hypothesis could serve as a general principle to improve inference efficiency in other domains as well. 

\section{Related Work}
\paragraph{Universal Morphology Control}
While generalization across robots with variations only in kinematics or dynamics parameters has been extensively studied in prior work \citep{tobin2017domain,peng2018sim,clavera2018learning,ghadirzadeh2021bayesian,feng2022genloco}, universal morphology control poses a more challenging task of generalization across morphologies. 
While learning an MLP policy with zero-padding is feasible to work across morphologies, it is significantly outperformed by GNN
\citep{wang2018nervenet,pathak2019learning,huang2020one} and TF
\citep{kurin2020my,dong2022low,furuta2022system,gupta2022metamorph,hong2022structureaware,trabucco2022anymorph,chen2023subequivariant,xiong2023universal}. 
Furthermore, TF outperforms GNN by better modeling the interactions between distant limbs in a robot \citep{kurin2020my}, thus has been adopted by most recent works. 
However, these more sophisticated models introduce higher computational cost during deployment, which we aim to tackle by generating a compact MLP controller at inference time with morphology-conditioned HN. 
Similar to our work, ModuMorph \citep{xiong2023universal} also utilizes HN. 
However, ModuMorph introduces HNs to better model the diverse behaviors across different limbs and only uses HNs to generate embedding and decoder layers inside a TF, while our method introduces HNs to improve inference efficiency via knowledge decoupling and generate a whole MLP controller with HNs. 

\paragraph{Hypernetworks}
A hypernetwork \citep{ha2016hypernetworks} takes context features as input to generate the parameters of a base network. 
In the context of RL, HNs provide a powerful way to model the complex dependency between task context and the optimal control policy for the task \citep{galanti2020modularity,sarafian2021recomposing}, and has been widely adopted in multi-task RL and meta-RL \citep{yu2019multi,peng2021linear,beck2022hypernetworks,rezaei2022hypernetworks,beck2023recurrent}. 
While these works mainly utilize the expressive power of HNs to improve task performance, we focus more on utilizing HN's knowledge decoupling ability to improve inference efficiency. 

\paragraph{Policy Distillation}
Distillation \citep{buciluǎ2006model,hinton2015distilling} transfers knowledge from a teacher model or multiple teacher models to a student model. 
In the context of RL, it has been used to compress the policy network \citep{rusu2015policy}, accelerate learning on a new task \citep{parisotto2015actor}, and facilitate policy learning \citep{lee2020learning,chen2022system,wan2023unidexgrasp++}. 
However, less attention is paid to measuring and minimizing the generalization gap between the teacher and the student. 
\citet{igl2020transient} distill a teacher policy into a student with identical architecture, which is used as an intermediate step to improve task generalization of a policy trained via RL. 
\citet{chen2022system} and \citet{wan2023unidexgrasp++} investigate generalization of a distilled policy to unseen objects (tasks) in robotic manipulation. 
However, as the teacher has access to privileged information while the student does not under their setting, there is always a performance gap between the teacher and the student, and this gap is larger on the test tasks than that on the training tasks, which indicates a generalization gap. 
Most similar to our work is \citet{furuta2022system}, which distill multiple single-robot MLP teachers into a TF student for universal morphology control. 
But they only show the advantage of using TF as the student architecture, but do not compare with a universal TF teacher to measure generalization gap. 
Furthermore, the TF student policy they learn still has efficiency issue during deployment. 
To the best of our knowledge, we are the first to systematically investigate how to mitigate generalization gap in policy distillation via proper algorithmic choices. 

\section{Conclusion}
In this paper, we propose \model to achieve both good performance and high efficiency at inference time for universal morphology control. 
While training \model via RL is hard, we take a policy distillation approach, and systematically investigate how some key algorithmic choices influence task generalization of the student policy. 
\model matches the performance of the universal TF teacher on both training and test robots, while significantly reducing memory and computational cost, which supports our hypothesis that decoupling inter-task and intra-task knowledge can improve inference efficiency. 

Given the wide application of TF in foundation models and their extremely high inference cost \citep{achiam2023gpt,brohan2022rt,brohan2023rt,padalkar2023open}, the findings in our paper could serve as a general principle to reduce the inference cost of TF in other domains as well. 
However, a potential limitation when extending our method to other domains is the additional cost introduced by the HN. 
This is not an issue for universal morphology control, as the morphology context remains unchanged on each robot. So we only need to call the HN once before deployment while the whole HN can be discarded afterwards.
However, if task context changes over time, such as controlling a robot to solve different tasks by following language instructions, the HN can not be discarded and needs to be called whenever a new instruction is given. This requires additional space to save the HN parameters, and the inference efficiency gain may decrease as the HN will be called more frequently. 
How to tackle this challenge is an interesting direction for future work. 

\section*{Impact Statement}
This paper presents work whose goal is to advance the field of Machine Learning. There are many potential societal consequences of our work, none which we feel must be specifically highlighted here. 

Specifically, as the control policy learned by our method can be deployed on real-world robots for tasks like locomotion and manipulation, issues like safety in operation need to be carefully considered, whose impacts have been extensively discussed in prior work. 

\section*{Acknowledgements}
We would like to thank Matthew Jackson and Alex Goldie for their helpful discussion on the work. 
We would also like to thank the conference reviewers for their constructive feedback on the paper. 
Zheng Xiong is supported by UK EPSRC CDT in Autonomous Intelligent Machines and Systems (grant number EP/S024050/1) and AWS.
Risto Vuorio is supported by EPSRC Doctoral Training Partnership Scholarship and Department of Computer Science Scholarship. 
Jacob Beck is supported by the Oxford-Google DeepMind Doctoral Scholarship. The experiments were made possible by a generous equipment grant from NVIDIA. 

\bibliography{reference}
\bibliographystyle{icml2024}

\newpage
\appendix
\onecolumn
\section{Morphology Context Features and Transformations}
\label{appendix:context-features}
For each robot, its morphology context includes the topology graph of the robot and limb-wise context features. 
Limb-wise context features include: (1) The initially relative position of the limb w.r.t.\ its parent node; (2) The initially geometric orientation of the limb w.r.t.\ its parent node; (3) The mass and shape parameters of the limb; (4) Parameters about the joints that connect the limb to its parent node, including joint type, joint range and axis, and motor gear \citep{gupta2022metamorph}. 

As discussed in Section \ref{subsec:architecture}, the context features need to be distinguishable enough to learn a good morphology-conditioned HN. 
Among the original context features, we find that the relative position feature does not provide sufficient discrimination, as different limbs, especially symmetric ones within a robot, can have the same relative position to their parents, but actually locate in different parts of the robot and play different roles. 
To solve this issue, we transform relative position features into absolute position features to better distinguish different limbs' positions in the morphology, which can be easily computed by following the path from the torso (the root of the morphology tree) to each limb node. 
Experimental results in Figure \ref{fig:HN-ablation} validate the importance of this feature transformation. 

\section{Further Experimental Setup}
\subsection{Generation of PD Robots}
\label{appendix:PD-robot-generation}
We mutate the 100 training robots in the UNIMAL benchmark to get an augmented set of 1000 PD robots, on which we collect expert data for policy distillation. 
Specifically, for each training robot, we first uniformly sample a number $m$ from $\left[ 1,2,3 \right]$, then sequentially apply $m$ mutation steps to the robot to get a new morphology. 
See \citet{gupta2021embodied} for the set of feasible mutations allowed in the UNIMAL benchmark. 
We generate 9 variants for each training robot, yielding an augmented set of 1000 PD robots in total. 

\subsection{FLOPs Computation}
\label{appendix:flops-computation}
We compute the FLOPs of a linear layer with input dimension $M$ and output dimension $N$ as $2 \times M \times N$ \citep{epoch2021backwardforwardFLOPratio}. 
Then the FLOPs of an MLP or a TF can be analytically computed by recursively decomposing its FLOPs into the summation of basic linear layers' FLOPs. 
We omit the FLOPs of other operations like activation functions, layer normalization and etc., as these operations only have linear complexity, which is negligible compared to the quadratic complexity of linear layers. 

\subsection{Architecture Details of Different Methods}
\label{appendix:model-size}
Table \ref{tab:model-size} shows the size of different architectures in each environment. 
For the ModuMorph (teacher) and ModuMorph (oracle), we use the same architecture hyperparameters as in \citet{xiong2023universal}. 
For the base MLP in \model, we fix the hidden layer size to 256, and do a grid search over hidden layer number to find the smallest base MLP that can achieve on-par performance with the teacher policy in each environment. 
Then we set multi-robot MLP's architecture identical to the base MLP in \model. 
For ModuMorph (compressed) and TF (compressed), we tune the number of attention layers, the number of attention heads, and the dimension of the linear layer's hidden dimension inside the attention module, so that the compressed model has a similar number of parameters as \model's base MLP. 
The token embedding dimension is 128 for all different TF architectures. 
\begin{table}[h]
    \centering
    \resizebox{\columnwidth}{!}{
    \begin{tabular}{c|cc|ccc|ccc|ccc}
        \toprule
        Environment & \multicolumn{2}{c|}{Base MLP in \model} & \multicolumn{3}{c|}{ModuMorph (teacher)} & \multicolumn{3}{c|}{ModuMorph (compressed)} & \multicolumn{3}{c}{TF (compressed)} \\
         & \multicolumn{2}{c|}{\& Multi-robot MLP} & \multicolumn{3}{c|}{\& ModuMorph (oracle)} & \multicolumn{3}{c|}{} & \\
         & Layer num. & Hidden size & Layer num. & Head & Hidden dim & Layer num. & Head & Hidden dim & Layer num. & Head & Hidden dim \\
         \midrule
        FT & 2 & 256 & 5 & 2 & 1024 & 1 & 1 & 128 & 1 & 1 & 256 \\
        VT & 3 & 256 & 5 & 2 & 1024 & 1 & 1 & 128 & 2 & 1 & 128 \\
        Obstacle & 3 & 256 & 5 & 2 & 1024 & 1 & 1 & 128 & 2 & 1 & 128 \\
        \bottomrule
    \end{tabular}
    }
    \caption{The size of different models in each environment.}
    \label{tab:model-size}
\end{table}

\section{Further Experimental Results and Analysis}

\subsection{Further Analysis on Inference Efficiency}
\label{appendix:efficiency-analysis}
As shown in Table \ref{tab:distill-result}, \model's efficiency advantage is more significant in the FT environment, as in the other two environments, there is an additional high-dimensional terrain information input to the policy, which needs to be processed by a large MLP encoder. 
This adds a large constant to the model size and FLOPs of all methods, and thus reduces the efficiency ratio of \model to the other methods.

Although multi-robot MLP and the base network of \model have the same number of hidden layers and hidden units, the model size and FLOPs of multi-robot MLP are still a bit larger than those of \model, as it needs to condition on the morphology context by concatenating limb-wise context features to the policy input, which introduces some additional cost. 

In \model, generating the base MLP with HN takes 43M FLOPs in the FT environment, and 65M FLOPs in the VT and Obstacle environment. The FLOPs of generating the base MLP with HN is just slightly larger than the FLOPs of a single inference step with a universal TF controller. 

\subsection{Ablation Study on Context Representation Learning}
\label{appendix:context-representation-ablation}
\begin{figure}[t]
    \centering
    \includegraphics[width=\columnwidth]{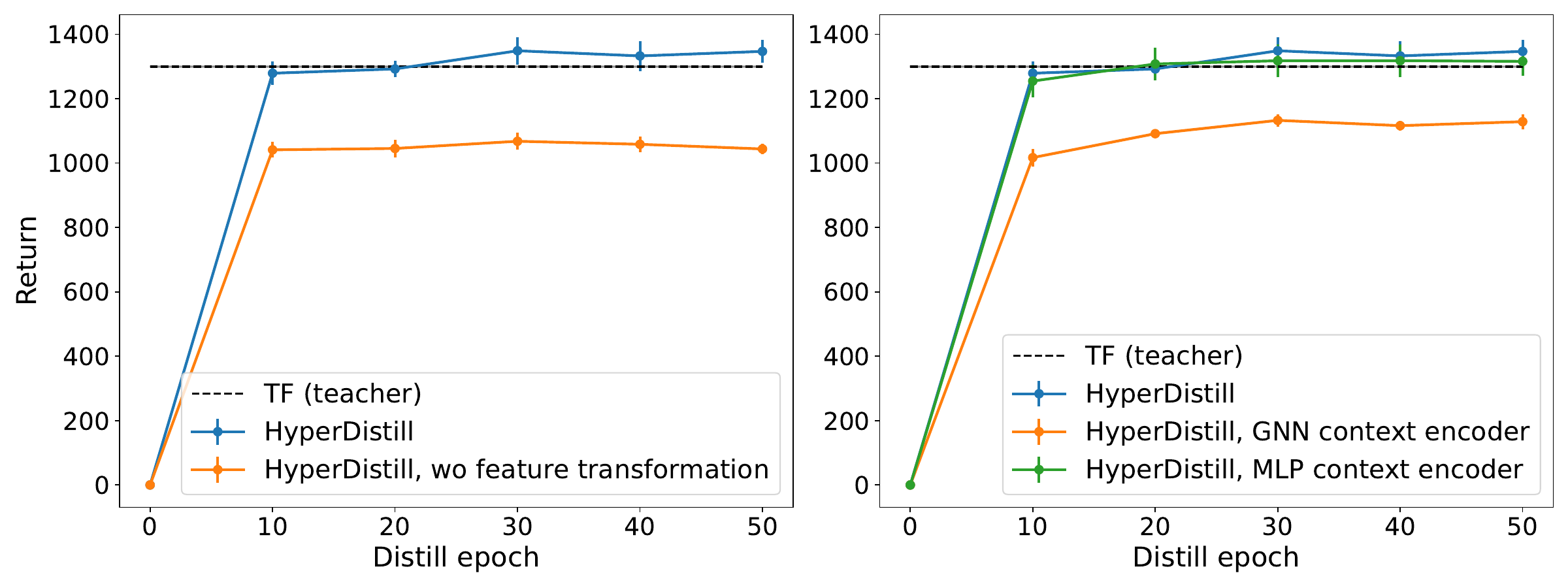}
    \caption{How context representation learning in \model influences generalization performance in FT. Left: context feature transformation. Right: architecture of the context encoder.}
    \label{fig:HN-ablation}
\end{figure}
\paragraph{Context Feature Transformation}
As shown in Figure \ref{fig:HN-ablation} (left), conducting feature transformations to improve feature discrimination plays an important role in improving performance. 

\begin{figure}[t]
    \centering
    \includegraphics[width=\columnwidth]{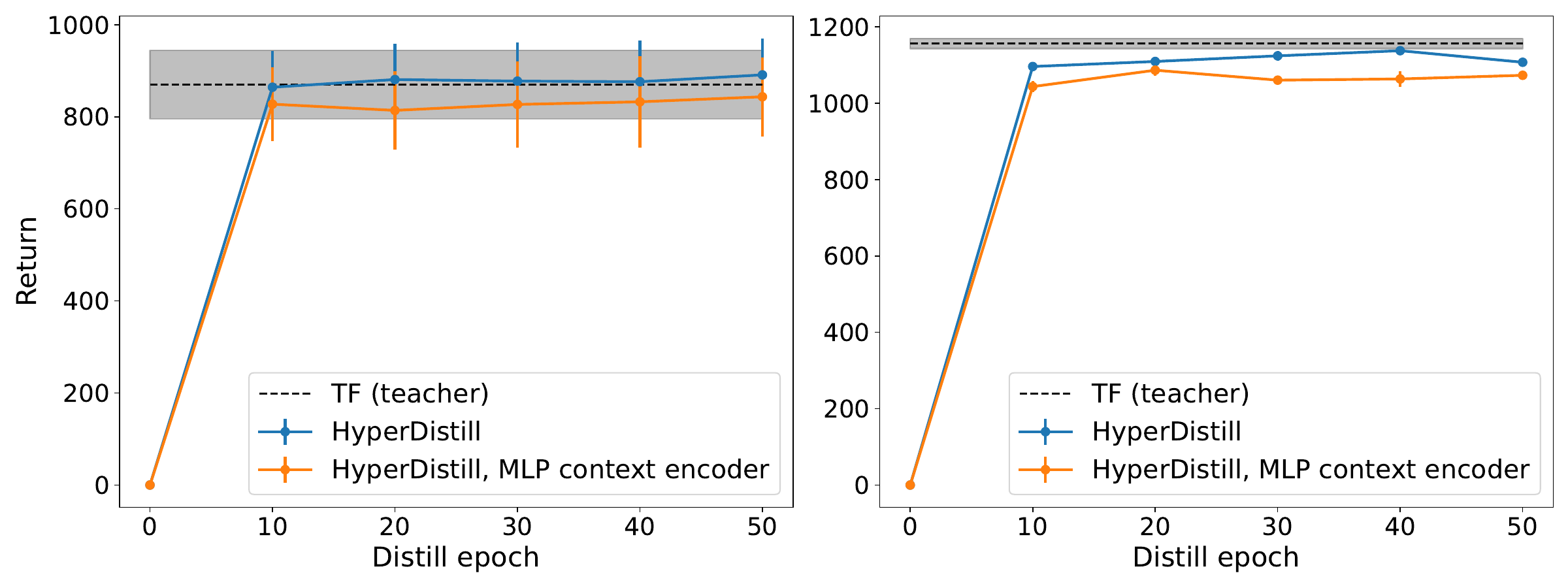}
    \caption{How TF and MLP context encoder influence \model's generalization performance. Left: VT; Right: Obstacle.}
    \label{fig:ablation_context_encoder}
\end{figure}

\paragraph{Architecture of Context Encoder}
As shown in Figure \ref{fig:HN-ablation} (right), for different choices of the context encoder, GNN performs the worst, possibly due to the over-smoothing issue of GNN \citep{chen2020measuring} which harms discrimination of the context embedding. 
While the MLP context encoder does not consider node interaction, it still achieves quite good performance in FT. 
We thus further compare MLP and TF context encoder in the other two more challenging environments (Figure \ref{fig:ablation_context_encoder}), and find that the TF context encoder performs better, which validates the benefits of modeling node interaction for context representation learning. 

\end{document}